\documentclass[abstracton]{scrartcl}
\usepackage[backend=biber, citestyle=numeric, bibstyle=numeric,sortcites]{biblatex}
\usepackage{graphicx}

\usepackage{amsmath, amsfonts, amssymb}      
\usepackage{setspace}

\newcommand{\x}[1]{\textit{#1}}
\newcommand{\y}[1]{\textbf{#1}}

\begin{document}

\title{An argument for the impossibility of machine intelligence}
\author{Jobst Landgrebe \and Barry Smith}
\date{\today}
\publishers{University at Buffalo\\
Department of Philosophy\\
              \texttt{\small jobstlan@buffalo.edu}           
              \texttt{\small phismith@buffalo.edu}           
}


\maketitle

\normalsize

\begin{abstract}
Since the noun phrase `artificial intelligence' (AI) was coined, it has been debated whether humans are able to create intelligence using technology.
We shed new light on this question from the point of view of themodynamics and mathematics.
First, we define what it is to be an agent (device) that could be the bearer of AI. Then we show that the mainstream definitions of `intelligence' proposed by Hutter and others and still accepted by the AI community are too weak even to capture what is involved when we ascribe intelligence to an insect.
We then summarize the highly useful definition of basic (arthropod) intelligence proposed by Rodney Brooks, and we identify the properties that an AI agent would need to possess in order to be the bearer of intelligence by this definition.
Finally, we show that, from the perspective of the disciplines needed to create such an agent, namely mathematics and physics, these properties are realizable by neither implicit nor explicit mathematical design nor by setting up an environment in which an AI could evolve spontaneously.

\bigskip

\noindent \textit{Keywords: Artificial Intelligence (AI),  Complex Systems, Ergodicity, Phase Space, Context-dependence}
\end{abstract}
\section{Introduction} \label{intro}

Since \textcite{turing:1951} and others initiated Artificial Intelligence (AI) research as a discipline in its own right in the late 1940s, there has been a constant debate about whether AI can be achieved at all.
\textcite{dreyfus:1972} was one of the first to fundamentally question the feasibility of AI based on his Heideggerian view of the human mind. At the opposite extreme there are those -- drawing on the incredible advances in AI in recent years in areas such as abstract games and machine translation -- who have defended the idea that machines might be created with an intelligence vastly superior to that of our own species \cite{chalmers:2010, bostrom:2014}.

The opinions expressed in this debate depend in every case on how `intelligence' is to be defined.
On the still standard view in the AI community, intelligence is defined as the ability of an agent to achieve a maximal reward sequence in a given environment at minimum computational cost \cite{hutter:2007}.
We show in \ref{defAI} that this and the other similar definitions proferred by the AI community are not only unable to capture human intelligence; they cannot capture even the intelligence manifested in insect behaviour.
This is because they fail to model perception, motion (as a sensorimotor activity), and the non-Markovian, erratic nature of reward sequences.

A more substantial definition of intelligence is therefore required. To arrive at such a definition, to assess its implications for the issue of machine intelligence, we proceed as follows.
First, we define what it is to be an agent that could be the bearer of AI.
Then, we introduce a widely accepted and highly useful definition of basic (arthropod) intelligence.
We then identify the properties that an AI agent would need to possess in order to be the bearer of intelligence by this definition.
And finally we show that, from the perspective of the disciplines needed to create such an agent, namely mathematics and physics, these properties are unrealizable. 
This means that, no matter what mathematical models and engineering methods we deploy, we will still not be able to create a Brooks-intelligent agent of the needed sort, and thus also will not be able to create an intelligence comparable to that of human beings. Such methods may, of course, be developed in the future, but only if the nature of these disciplines is fundamentally transformed.

If we cannot engineer an intelligent agent, might it be possible to create the conditions under which an AI would emerge spontaneously -- an idea advanced, for example, by Chalmers? In a coda, we show that this idea, too, is not supportable.
We conclude that humans are unable to engineer AI agents, and that they are also unable to create a sitution in which such agents could emerge spontaneously.
Thus, there will be no intelligent machine in the sense of our proposed definition of intelligence, not even intelligence of the sort possessed by bees and other arthropods.
It follows that higher intelligence (mammalian or human) is unfeasible also.

\section{Engineering Intelligence?}

\subsection{Definition of `agent' and of `Artificial Intelligence'}\label{defAI}

We define an agent as a man-made technical artefact that can perform actions in an open environment based on the events it senses in this environment.
Thus it can perceive and move in ways that take account of what it perceives.
For example, a self-driving car would be an agent, while an assembly-line robot or a confined self-driving train (such as are to be found in many airports) is not.

An intelligent agent must be able to  interact with an open environment, namely the world that surrounds us.
Any definition of intelligence that would not require this ability is insufficient.
An agent unable to cope with natural environments would be less intelligent than the most primitive organisms.
The AI-robotics pioneer Rodney Brooks recognized this already 30 years ago when he described his goal of building robots that can `move around in dynamic environments, sensing the surroundings to a degree sufficient to achieve the necessary maintenance of life and reproduction.' \cite[140]{brooks:1991}

The \x{utility-based} definitions proposed by mainstream AI authors \cite{hutter:2007, schmidhuber:2007, pennachin:2007, yudkowsky:2004, muehlhauser:2012a} cannot meet this requirement.
The utility function they provide\footnote{The definition displayed here is from \textcite{hutter:2007}, who refers to it as a definition of `universal intelligence'. The other cited definitions are isomorphic thereto. A more detailed statement of the mathematical and other problems associated with these definitions is provided in
\cite{landgrebeSmith:2022}}:

\begin{equation}\label{eq:eqV}
V_{\mu}^{\pi} := \mathbf{E}(\sum_{i=1}^{\infty}r_i) \leq 1
\end{equation}

\noindent defines the utility $V_{\mu}$ of an agent $\pi$ for an environment $\mu$ (where by environment is intended the corresponding minimal binary \textit{description}), as the stochastic expectation $\mathbf{E}$ over the infinitesimal series of the rewards $r_i$ that the agent obtains as a result of its behaviour at each step. 

The definition limits the maximum utility to $1$ and implicitly ensures that simple environments which can be solved in a few steps will not receive a high utility score. Unfortunately it faces two main problems: (i) it is unable to take account of the active, iterative and directed, intentional nature of perception \cite{merleau:1945, gibson:1979}. 

Thus, for example, it cannot take account of what occurs when a predator actively observes the behaviour of its prey and follows it through changing environments.
The static variable $\mu$ captures none of this.
(ii) Its step-by-step, additive reward scheme cannot account for the sorts of reward patterns an agent can encounter in a real-world environment.
Indeed, it is not possible to define on its basis even a reward sequence that mimics the reward pattern experienced by a natural intelligence such as a fruitfly, an aspect to which we return below.

On the basis of equation (1) Hutter (and the other authors listed) provide a definition of intelligence $\Upsilon$
as the sum of complexity-penalised\footnote{The penalisation is achieved via the Kolmogorov complexity function $K$, which indicates the length of the shortest algorithm needed to computationally represent the environment description $\mu$ of the agent $\pi$.} utilities an agent can achieve over a set $U$ of environments defined for the agent:

\begin{equation} \label{eq:Yps}
\Upsilon(\pi) := \sum_{\mu \in \mathbf{U}} 2^{-K(\mu)} V_{\mu}^{\pi}
\end{equation}

where $U$ is the set of environment descriptions with which the agent can cope, and $V_{\mu}^{\pi}$ is the utility achieved by the agent $\pi$ for a given environment description $\mu \in U$.
For AlphaZero, for example, $U$ is the set of environments (board positions, opponent activities, etc.) occurring in games it can play.
For a putative universal agent, of course, all possible environments would have to be taken into account \cite[section~3.1]{landgrebeSmith:2022}.
Though the infinitesimal definition of utility in (1) and the penalisation of complexity in the definition of $\Upsilon$ provide a statistically robust measure of the kind of surrogate intelligence those working in the general artificial intelligence (AGI) field have decided to focus on, the definition is too weak to describe or specify the behaviour even of an arthropod. 
This is not only obvious from the issues already mentioned above, but also from the fact that algorithms which realise the reward-schemes proposed in (1) and (2) (for example, neural networks optimised with reinforcement learning) fail to display the type of generalisable adaptive behaviour to natural environments that arthropods are capable of, for example when ants or termites colonise a house.

\subsection{What an AI agent needs}

The definition given by Brooks, on the other hand, applies directly to natural environments.
It is useful for our purposes, first, because it circumvents the underspecification of intelligence of the type shown in Hutter-style equations, which fails to model the way intelligent organisms interact with nature. 
But it is also useful because it avoids the thorny issues involved in providing a definition precisely of human-level intelligence.
As Brooks saw, a machine at the intelligence level of an ant with a goal system that could be parameterised by humans would already be highly useful.

Brooks defines an AI agent, again, as an artefact that is able `to move around in dynamic environments, sensing the surroundings to a degree sufficient to achieve the necessary maintenance of life and reproduction'. 
The natural environment of a moth, for example, is a dynamic environment in the relevant sense, and so also are the hybrids of natural environment and man-made technology that we humans encounter on a city street or highway, in a shopping mall or nuclear reactor, or while flying a helicopter in a mountain valley.
Such environments are `dominated' by complex systems in the sense defined in thermodynamics \cite{thurner:2018}, where a `system' is defined as a totality of dynamically interrelated physical elements. To delimit a system is to select a level of granularity of its elements, from microphysical particles to entire galaxies, and to identify the types of interactions between such elements as the system exists through time.

All living organisms are complex systems \cite{prigogine:1955, prigogine:1973, mora:1963}, but so also are many inanimate systems, for example the Earth's weather system, climate system, and geological water cycle \cite{thurner:2018}. 
Such systems share many features. 
Here we will examine only those features needed to show that an agent able to  interact intelligently with such a system cannot -- using the mathematics which exists today -- be engineered. 
First, however, we need to contrast complex systems with what we shall call `logic systems', which are systems modelled, designed and built using extended Newtonian mathematics, including the standard logical inference machinery.

\subsubsection{Some properties of logic systems}
All technical artefacts we build are logic systems in this sense, and in exercising their main functions their behaviour is deterministic.\footnote{Those which consume energy have a dynamic aspect which Prigogine called `dissipative' (because the energy they consume is dissipated in the form of some kind of motion, radiation or heat). This is true, too, of your phone and your laptop, but it is an aspect that is in most situations not relevant for main their function or to our argument here.} 

There are some who hold that engineers often build technical artefacts not on the basis of mathematical models but rather by using trial and error, followed by physical testing of the results.
There are indeed some who hold that something similar is the case when engineers build digital neural networks and other, similar stochastic artefacts.
There may indeed by some engineered devices which we built in this way.
But if they realize their intended functions then there is (or can be built) in every case a mathematical model which represents how they work with an approximation close to the measurement error. 

In the case of neural networks the matter is still clearer.
Neural networks are indeed built by a process of training, which means that they are fed with sample data selected as material for the algorithm to learn from.
But neural networks are logic systems.
They are engineered and trained to run on computers, which means that each one of them can of necessity be identified with some algorithm.
This may be, highly complex -- in the case of the GTP-3 algorithm, for example, it consists of some 
175 billion parameters.\footnote{It is this merely algorithmic complexity which explains why the search for `explainable AI' must often remain fruitless.}

There are three additional properties of logic systems of importance for our argument here: 

\begin{enumerate}
\item Their phase space is fixed.
\item Their behaviour is ergodic with regard to their main functional properties.
\item Their behavior is to a large extent context-independent. 
\end{enumerate}

What does this mean?

\paragraph{Phase space} A system can be mathematically characterised by variables describing the behaviour its elements. These variables are called `state variables' in physics.
The phase space of a system is the vector space whose basis is formed by the state variables relating to each element type of the system.
This space has as many dimensions as the number of distinct state variables that relevant elements of the space can have.
The variables in this space are determined by the way the system is modelled mathematically, and thus in any given case it consists of a finite set of measurable values.
A fixed phase space may allow almost exact\footnote{`Almost exact' means that measurements performed on a system will deviate only from the model in the order of magnitude of the measurement error. 
For example, the basic Newtonian model of the earth's orbit around the sun based on the gravitational interaction of these two bodies is almost exact, but does not take into account the influence of other bodies, which lead to slight periodic deviations of the earth from its orbit.} mathematical modelling of a system, for example using differential equations.

\paragraph{Ergodicity} 
Ergodicity describes the stochastic behaviour of a system.
A system is called ergodic if, over sufficiently long periods of time, the time in which a system element occupies any given region of the phase space is proportional to the volume of this region.
Access to the regions of the phase space is thus equiprobable over time.
For example, a steam engine, which is a logic system when account is taken only of its behaviour in exercising its main function,\footnote{The steam engine is a complex system when account is taken of the energy dissipation occurring in its steam boiler.} the elements (for example the pistons in the engine's cylinders) access all regions of the phase space in an equiprobable manner. 
This behaviour of the engine is deterministic and the speed of a locomotive driven by the engine can therefore be regulated via the pressure in its boiler.

\paragraph{Context-independence} 
A logic system is self-contained.
Its elements interact in the same manner independently of the system's environment as long as this environment does not alter the phase space of the system.
For example, the solar system as a gravitational system is a logic system, and its behaviour
would not be altered if it was moved some light weeks closer to the binary star system Alpha Centauri (which is 4.37 light years from the sun).\footnote{Things would be different if the sun came so close to Alpha Centauri that they would gravitationally interact in a non-negligible way.} In the same way, a motor vehicle is a logic system that can be used on any terrain for which it was built, no matter where it is -- as long as there is gravitation, an approximatively even ground, a slope not surpassing a certain angle, and oxygen in the air for its combustion engine.

Note that the goal of AI research, in our terms, is to build a logic system  that is at least Brooks-intelligent.

\subsubsection{Some properties of complex systems}
The systems that make up our environment and interact with each other are not logic systems but complex systems.
Our world is shaped by complex systems, some of them inanimate, such as the tectonic movements caused by the Earth's seismic system; and some of them animate, in particular all the behaviour of humans and other animals. As we learn from we standard mathematical theory of complex systems \cite{thurner:2018}, all such systems, including the systems of complex systems resulting from their interaction,  
\begin{enumerate}
\item have a variable phase space,
\item are non-ergodic, and
\item are context-dependent.
\end{enumerate}
\noindent (1.) A \y{variable phase space} means that the variables which define the elements of a complex system can change over time.
New elements and new associated variables can arise, and others can disappear.
Existing elements can obtain new variables, or in other words new dimensions along which their behaviour may vary , as occurs for intsance when a child learns to walk, or to swim, or to play tag.
The way the elements of the system interact can change.
New types of interaction can arise.
All this applies, for example, to the way people evolve over their lives, to the interactions of animals in groups, or to the way conversations (and indeed entire languages) evolve over time.

\noindent (2.) A \y{non-ergodic} system produces erratic distributions of its elements. No matter how long the system is observed, no laws can be deduced from observing its elements. A simple example is a group of children in a school playground. Though patterns emerge, they can change at any time and for any one of an open-ended number of reasons.

\noindent (3.) A \y{context-dependent} system changes its elements and the way they interact depending on its environment. Examples are (i) the total dependence of those species which are specialized to specific environments upon those environments (which means that they perish in other environments), and (ii) the adaptation of certain species to multiple environments which lets them change their behaviour when they switch from one environment to another.

\subsection{Why we cannot model, design and build an agent that is Brooks-intelligent}
Each AI agent, by our definition of `agent', will have to cope with a complex-system-generated environment.
An AI agent is a technical entity and needs to be modelled, designed and built (engineered).
Modelling such an agent means to define its functions and the way it is supposed to interact with its environment.
Designing it means to specify how these functions are to be realised physically.
Both steps require mathematical models.

There are two fundamental types of mathematical model: deterministic and stochastic (along with hybrids, usually obtained by composition).
Any mathematical model requires a vector space, often a coordinate space such as the Cartesian space $\mathbb{R}^2$ or the Euclidean space $\mathbb{R}^3$ over an algebraic space such as that formed by the real numbers $\mathbb{R}$. 
Such a space is needed to define the variables used in the mathematical equations which are supposed to model the corresponding real system.

\paragraph{Variable phase space}
But with the \x{changing variables and interactions} that we find in complex systems, there is \x{no coordinate space} over which models can be defined.
Any model is defined for a specific vector space, and is invalid if the reality targeted by the model differs from the vector space for which it was defined.
The more it differs, the stronger is the deviation, and the less accurate the model becomes.
With complex systems, the deviation can become huge, and this can happen suddenly, when new elements evolve.\footnote{This is true even though partial aspects of the system can be modelled approximatively. For example, we have an approximative, descriptive understanding of the physiology of blood pressure. Yet we do not understand essential hypertension and have no causal way to treat it once it emerges (unpredictably)\cite{jordan:2018}.}

Examples of such deviation, which arise due to the massive discrepancy between model and system, are reflected in the predictions generated in a weather forecast (which are never accurate over longer periods), climate models (which are never accurate) or epidemiological models (which are in extreme situations only approximatively accurate).
All these deviations become larger as time passes, because the effects which prevent exact models for complex systems gain importance exponentially \cite{thurner:2018}.

\paragraph{Non-ergodic process}
Yet more obstacles to modelling are created where we are dealing with \x{non-ergodic processes}, which produce events in which we cannot identify \textit{any} law-like pattern that could be modeled mathematically.
Of course, there are many patterns in complex systems which can be modelled. These include for example the different sorts of cyclic behaviour exhibited by all animals along multiple dimensions (respiration, heart beat, wake-sleep-cycle, feed-rest-hunt cycle, cycles of sexual behaviour, etc.). But these cover only partial processes and cannot be described using differential equations or other exact modelling approaches.
For example, we cannot even exactly model classical conditioning in a way that would be valid across multiple laboratories experimenting on harnessed arthropods (honey bees) even where the exact same experimental setup is being applied \cite{bitterman:2006}.

Importantly, the traces of non-ergodic processes (which means the series of data which such processes generate) provide no adequate target spaces for stochastic sampling, because the latter requires in every case that samples are produced from a distribution belonging to the exponential family or to a non-parametric distribution that can      be reliably estimated \cite{klenke:2013}.
The samples drawn from such traces are never representative of the process behaviour due to the non-ergodic character of the process.
There is no distribution family to sample from.
This systematically prevents stochastic modelling of such processes.
It is also the reason why reward traces like those described in equation (1) cannot be provided for the training of artificial agents where interactions with complex systems are involved.

There is no way to set up such a trace because there is no stochastic process available to create it.
A deterministic model would not work either because such a model, too, cannot produce the pattern of rewards produced by a complex system.

This is the reason why, for example, machines cannot learn to conduct real conversations using machine learning -- which is a method to obtain mathematical models implicitly using optimization methods applied to training data \cite{hastie:2008, goodfellow:2016, bertsekas:2016}.
The training data obtained from recorded conversations are never adequate as material to train a model because the conversations are conducted by complex systems.
If such data are used to train a stochastic model, the model will always be fitted to non-representative data.
It will therefore fail, if not in the next conversation, then in some conversation in the near future, and not in the way that humans fail regularly in conversations, but rather in a way that betrays that the model is not able at all to meet basic human expectations for a conversation with another human being \cite{landgrebeSmith:2021}.\footnote{Willingness to chat with a bot for entertainment purposes, a habit quite popular in some cultures, is a different matter.} 

Fully deterministic conversation agents are improving every day even as we speak.
But more than fifty years after the first chatbots appeared, they seem still not to have reached the level where they can successively and smoothly book or a room or a flight without requiring extra effort on the part of the human being involved.
The goal of a conversation agent that can make the sorts of spontaneous responses over a longer period that are characteristic of human beings is unreachable. 

\paragraph{Context-dependence} This property of complex systems has the consequence that the system will use a different phase space following different principles depending on the context in which it is situated.
A housefly cannot adapt to a window pane.
Its behaviour becomes unpredictable and erratic when it gets caught behind a window.
On the other hand, those animals that can adapt to new types of situations show a completely erratic pattern during the process of adaptation.
Because, in natural environments (as in financial markets), the context can radically change at any time, context dependence makes the behaviour of complex systems in such environments unpredictable. Of course, humans have developed strategies to make each other's behaviour more predictable in order to enable life in communities (based on mutual trust) and societies (based on social norms). But the omnipresence of conflicts (private and public, peacefully resolved or leading to violence) shows that predictability is still very low, since the interaction of complex systems is creating novel, unexpected situations all the time.

\subsubsection{The self-driving car amidst complex systems}

To summarise, an artificial agent, to emulate human-level intelligence, would have to cope with an environment arising from the interactions of animate and inanimate complex systems. But we have no way of modelling such interaction mathematically. Neither explicitly, using manually crafted models, nor implicitly, using stochastic models.\footnote{However, we can model partial aspects very successfully using AI \cite{landgrebeSmith:2021}.}

When an AI agent is interacting with simple systems, for example an agent built for purposes of missile defense, the system works by modelling the motion of the incoming missiles, which are simple systems, using Newton's Laws. The phase space of the missiles is fixed, their behaviour is Markovian, their trajectory is ergodic. 

When, in contrast, the AI agent is in the situation of a self-driving car (agent) moving through the dense traffic in downtown San Francisco or through the mountain passes surrounding Lake Tahoe two hundred miles to the North, then in both situations, the variables it may be confronted with variables which are permanently changing.
Although for the untrained eye of a non-mathematician, the car simply moves in Euclidean space over time, when viewed through the eyes on an engineer, it faces a huge phase space determined by the various entities crossing its path.
Because both animate and inanimate sytems often show erratic behaviour, new phase space dimensions not present in the training data can arise any time, for example if a flock of birds flies over a road at low altitude while a child is running onto the road to catch its ball; or a demented pedestrian stumbles accross the road in a dense fog massively diffracting the car's lights.
The total situation the car is facing is determined by a complex system of complex systems whose behaviour is at every level non-ergodic. We are unable to sample adequate training distributions from the traces created by such processes, and unable also to define an adequate reward-path to train a utility-based agent.

\subsection{Coda: Why AI agents cannot arise spontaneously}
Could we create a system in which intelligence could evolve spontaneously, as proposed by \textcite[p.~16]{chalmers:2010}? His argument runs as follows:

\begin{enumerate}
 \item Evolution produced human-level intelligence. 
 \item If evolution produced human-level intelligence, then we can produce AI 
(before long).
 
\line(1,0){150}

\item Absent defeaters, there will be AI (before long).
\end{enumerate}

The first premise is of course true.
But the second is false, because it implies an equivalence between evolution and human activity, an equivalence which does not obtain.
Intelligence arose by natural selection through interactions of our ancestors with the inanimate environment and with competing organisms.
There was no \textit{plan} to yield intelligence on the basis of which these processes occurred, but rather only the \textit{drive} to survive and reproduce that is built into every living organism.
This drive led to the rise of intelligence through an immense series of processes, almost all of which occurred by chance.

But we neither know how to engineer the drive that is built into all animate complex systems, nor do we know how to mimic evolutionary pressure, which we do not understand and cannot model (outside highly artificial conditions such as a Petri dish).
In fact, if we already knew how to emulate evolution, we would in any case not need to do this in order to create intelligent life, because the complexity level of intelligent life is lower than that of evolution.
This means that emulating intelligence would be much easier than emulating evolution \textit{en bloc}.

Chalmers is, therefore, wrong. We cannot engineer the conditions for a spontaneous evolution of intelligence.

\section{Conclusions}
As we saw earlier, a machine at the intelligence level of an ant with a goal system that could be parameterised by humans would be highly useful. But if we cannot create artificial intelligence even at this level, this means that we cannot create any artificial agent that deserves to be called `intelligent', since Brooks' definition targets intelligence of a very much weaker form. 

This has two consequences:

\begin{enumerate}
 \item We have no reason to fear artificial superintelligence \`a la \textcite{bostrom:2014}.
 \item We can still engineer very useful `narrow AI'.
\end{enumerate}

To the first point.
Given that we have chosen a very weak definition of intelligence to prove that even this cannot be engineered, we are also on the safe side regarding the question whether humans might one day be overrun by an AI of any sort.
For if we cannot engineer an arthropod level AI, then we cannot engineer, either, an AI that could initiate a process that would lead to AIs that could surpass human intelligence.

To the second point.
The advances in applied statistical learning since the late 1990s have been very impressive.
We can now solve many specific problems identified by humans using machine learning or a combination of machine learning with other algorithms.
We will see many very useful applications of this technology over the next decades, not only in the design and supervision of ever more sophisticated logic system devices, but also in machine interactions with animate complex systems.
However, these solutions will be architected by humans and dedicated to solving special problems, either using explicitly designed models or using optimisation procedures applied to training data.
The utility of the latter depend always on the quality and scope of the training data, trained functionals or operators cannot cope with data points which do not or only rarely occur in the training set.
They do not generalise, despite recent progress with stochastic `foundation models' \cite{bommasani:2021}. 
Therefore, they will always be restricted to a small scope of narrow AI, even if they can be adjusted to various such scopes with little training effort. But as the scope increases, the non-ergodic nature of the sampling distribution becomes tangible in the training data and the models become overfitted to samples that differ from reality. 

 \section*{Conflict of interest}
 The authors declare that they have no conflict of interest.

\printbibliography

\end{document}